\documentclass[conference]{IEEEtran}
\IEEEoverridecommandlockouts
\usepackage{cite}
\usepackage{amsmath,amssymb,amsfonts}
\usepackage{algorithm}
\usepackage{algpseudocode}
\usepackage{graphicx}
\usepackage{textcomp}
\usepackage{xcolor}
\usepackage{subcaption}
\usepackage{xurl}
\def\BibTeX{{\rm B\kern-.05em{\sc i\kern-.025em b}\kern-.08em
    T\kern-.1667em\lower.7ex\hbox{E}\kern-.125emX}}
\begin{document}

\title{Large Language Models as Particle Swarm Optimizers}

% \author{\IEEEauthorblockN{Anonymous Authors}}

\author{\IEEEauthorblockN{1\textsuperscript{st} Yamato Shinohara}
    \IEEEauthorblockA{\textit{Graduate School of Information Science and Technology} \\
        \textit{The University of Tokyo}\\
        Tokyo, Japan \\
        yamato-shinohara@g.ecc.u-tokyo.ac.jp}
    \and
    \IEEEauthorblockN{2\textsuperscript{nd} Jinglue Xu}
    \IEEEauthorblockA{\textit{Graduate School of Information Science and Technology} \\
        \textit{The University of Tokyo}\\
        Tokyo, Japan \\
        jingluexu@gmail.com}
    \and
    \IEEEauthorblockN{3\textsuperscript{rd} Tianshui Li}
    \IEEEauthorblockA{\textit{Graduate School of Information Science and Technology} \\
        \textit{The University of Tokyo}\\
        Tokyo, Japan \\
        tianshui@g.ecc.u-tokyo.ac.jp}
    \and
    \IEEEauthorblockN{4\textsuperscript{th} Hitoshi Iba}
    \IEEEauthorblockA{\textit{Graduate School of Information Science and Technology} \\
        \textit{The University of Tokyo}\\
        Tokyo, Japan \\
        iba@iba.t.u-tokyo.ac.jp}
    \and
    % \IEEEauthorblockN{5\textsuperscript{th} Given Name Surname}
    % \IEEEauthorblockA{\textit{dept. name of organization (of Aff.)} \\
    % \textit{name of organization (of Aff.)}\\
    % City, Country \\
    % email address or ORCID}
    % \and
    % \IEEEauthorblockN{6\textsuperscript{th} Given Name Surname}
    % \IEEEauthorblockA{\textit{dept. name of organization (of Aff.)} \\
    % \textit{name of organization (of Aff.)}\\
    % City, Country \\
    % email address or ORCID}
}

\maketitle

\begin{abstract}
    Optimization problems often require domain-specific expertise to design problem-dependent methodologies.
    Recently, several approaches have gained attention by integrating large language models (LLMs) into genetic algorithms.
    Building on this trend, we introduce Language Model Particle Swarm Optimization (LMPSO), a novel method that incorporates an LLM into the swarm intelligence framework of Particle Swarm Optimization (PSO).
    In LMPSO, the velocity of each particle is represented as a prompt that generates the next candidate solution, leveraging the capabilities of an LLM to produce solutions in accordance with the PSO paradigm.
    This integration enables an LLM-driven search process that adheres to the foundational principles of PSO.
    The proposed LMPSO approach is evaluated across multiple problem domains, including the Traveling Salesman Problem (TSP), heuristic improvement for TSP, and symbolic regression.
    These problems are traditionally challenging for standard PSO due to the structured nature of their solutions.
    Experimental results demonstrate that LMPSO is particularly effective for solving problems where solutions are represented as structured sequences, such as mathematical expressions or programmatic constructs.
    By incorporating LLMs into the PSO framework, LMPSO establishes a new direction in swarm intelligence research.
    This method not only broadens the applicability of PSO to previously intractable problems but also showcases the potential of LLMs in addressing complex optimization challenges.
\end{abstract}

\section{Introduction}

Large Language Models (LLMs) have recently gathered attention for their capacity to directly address optimization tasks by formulating problems in natural language and utilizing the LLM’s reasoning abilities to generate candidate solutions~\cite{yang2024large, liu2024large, nie2024importance, zhou2023large}.
Compared to traditional techniques that demand hand-engineered operators or domain-specific expertise, LLM-based approaches can flexibly encode search strategies, constraints, and heuristics using prompts, thereby offering a more generalizable mechanism for controlling the optimization process.

Although such methods have demonstrated promise, prior work often integrates LLMs into existing frameworks without explicitly leveraging \textit{swarm intelligence} algorithms.
The swarm intelligence methods naturally lend themselves to multi-agent coordination and can benefit from LLM-enhanced communication or decision-making~\cite{wu2024autogen}.
In particular, Particle Swarm Optimization (PSO)\cite{kennedy1995particle, shi1998modified}  offers a simple yet powerful foundation wherein particles cooperate and learn from each other in a shared search space.
This simplicity, alongside its agent-based design, makes PSO an attractive basis for investigating how LLMs might guide or generate solutions in a collective, multi-agent setting.

In this study, we introduce \textit{Language Model Particle Swarm Optimization (LMPSO)}, a novel approach that explicitly incorporates an LLM into PSO by treating each particle’s update step as a prompt-driven process.
Rather than employing problem-specific update formulas, LMPSO constructs structured prompts—which encode velocity, position, and other contextual information—and feeds them to an LLM to obtain new candidate solutions.
This design extends PSO into a hyper-heuristic framework, where an LLM can dynamically generate or refine heuristics without requiring extensive hand-crafted operators~\cite{burke2013hyper,gendreau2010handbook}.

One key advantage of LLM-based optimization is its capacity to handle solution representations that may be cumbersome or infeasible in standard PSO pipelines.
For instance, while PSO was originally designed for continuous optimization problems, applying it to combinatorial problems like the Traveling Salesman Problem (TSP) has traditionally required developing velocities, directions between particles’ positions, and tailored update rules~\cite{wang2003particle}.
LMPSO simplifies this process by using natural language prompts, eliminating the need for manually crafted operators and encoding schemes.

Additionally, LMPSO functions as a hyper-heuristic, enabling the improvement of existing heuristics through a prompt-driven search, as demonstrated in prior research on evolving heuristic strategies~\cite{chen2024evoprompting, Romera-Paredes:2024aa}.
Also, LLM’s ability to understand contextual information about the given problem can be utilized to generate effective solutions, particularly in problems with rich data like symbolic regression.

Our goal is not to compare solution diversity with existing LLM-based optimizers, but to highlight how explicitly incorporating an LLM into a multi-agent, swarm intelligence framework can expand the range of solvable problems and the forms of final solutions.
Through experiments on both combinatorial and natural-language-based tasks, we illustrate LMPSO’s effectiveness in addressing problems that demand flexible or context-rich representations.

\section{Related Works}

Leveraging Large Language Models (LLMs) with advanced contextual understanding and reasoning capabilities for direct optimization has recently attracted considerable attention. In \textit{Optimization by PROmpting (OPRO)}, Yang et al.\cite{yang2024large} demonstrated that optimization problems can be effectively solved simply by describing them in natural language, then generating new solutions via an LLM based on previously obtained solutions. Along a similar line, \textit{AgentHPO}\cite{liu2024hyper} proposed a hyperparameter optimization framework that employs an LLM to automate the tuning process, providing an efficient and interpretable alternative to traditional AutoML approaches. Beyond these examples, LLM-based methods have shown promise in a variety of domains, including heuristic generation~\cite{Romera-Paredes:2024aa,ye2024reevo,liu2024evolution}, mathematical tasks~\cite{Romera-Paredes:2024aa}, and mixed-integer linear programming~\cite{ahmaditeshnizi2024optimus}.

A representative example of integrating LLMs into evolutionary algorithms is LLM-driven EA (LMEA)\cite{liu2024large}. By incorporating instructions for parent selection, crossover, and mutation within a “meta-prompt,” LMEA is able to generate new individuals and explore the global optimum. Experimental results suggest that LMEA performs comparably to existing methods on the Traveling Salesman Problem (TSP) with up to 20 cities, and even surpasses OPRO in some instances. One key advantage of these LLM-driven approaches lies in their ability to describe optimization problems in natural language, thereby lowering the barrier for problem-specific expertise compared to conventional methods. This capability broadens the range of problems for which LLM-based optimization techniques are applicable\cite{guo2023towards,pluhacek2023leveraging}.

In \textit{Large Language Models as Evolution Strategies}\cite{10.1145/3638530.3654238}, researchers showcased zero-shot optimization in a black-box setting using LLMs. Moreover, in \textit{The Importance of Directional Feedback for LLM-Based Optimizers}\cite{nie2024importance}, the authors illustrated that providing directional feedback within prompts enables LLMs to tackle diverse tasks, such as maximizing mathematical functions and composing poems. Collectively, these findings underscore how the content and structure of prompts significantly affect the performance of LLM-based optimizers.

Despite the promising advances of LLM-assisted approaches across a wide range of optimization tasks, further investigation is required to systematically integrate LLMs into established optimization paradigms. In particular, swarm intelligence algorithms present a compelling opportunity due to their multi-agent foundations—an area in which LLMs have already demonstrated substantial potential~\cite{guo2024large, chan2023chateval}. Integrating LLMs into swarm-based methods could thus unlock new possibilities for addressing complex, distributed problems that rely on collective agent-based reasoning and search capabilities.

\section{Proposed Method}

In this study, we propose \textit{Language Model Particle Swarm Optimization (LMPSO)}, which leverages an interactive LLM to update the position of each particle in Particle Swarm Optimization.

\subsection{Algorithm Overview}

LMPSO, as shown in Algorithm~\ref{alg::lmpso}, extends standard PSO by using the LLM to update solutions through prompts that define each particle’s velocity.
The optimization process begins with the initialization of particle positions, generated either randomly or by the LLM, and velocities represented as prompts like “Generate a position randomly” (lines~3–4).
The main loop iterates for up to $G$ iterations (line~5), where each particle’s objective value $f(x_{i}^t)$ is evaluated to update the personal best $pbest_i$ and global best $gbest$ (lines~7–12).

In each iteration, a new velocity $v_{i}^{t+1}$ is constructed based on the problem $T$, $pbest_i$, and $gbest$ (line~13), and a meta-prompt incorporating $T$, $v_{i}^{t}$, $x_{i}^{t}$, and $v_{i}^{t+1}$ guides the LLM to generate a candidate solution $x{\prime}_{i}$ (lines~14–15).
If $x{\prime}_{i}$ satisfies constraints, it becomes the new position; otherwise, retries or random reinitialization maintain diversity (lines~16–19).
This process continues until the maximum iterations $G$, after which the global best $gbest$ is returned as the optimal solution (line~23).
\begin{figure}[!t]
    \centering
    \begin{algorithm}[H]
        \caption{Language Model Particle Swarm Optimization}
        \label{alg::lmpso}
        \begin{algorithmic}[1]
            \State \textbf{Input:} Optimization Problem $T$, Objective function $f$, number of particles $N$, maximum iterations $G$
            \State \textbf{Output:} The best solution $g_{\text{best}}$
            \State Initialize the positions $x_{i}^{t}$ of $N$ particles randomly within the search space
            \State Initialize the velocities $v_{i}^{t}$ of $N$ particles with a prompt like "Generate a position randomly"
            \For{$t = 1$ to $G$}
            \For{each particle $i$ in the swarm}
            \If{$f(x_{i}^{t})$ is better than $f(pbest_i)$}
            \State Update $pbest_i \gets x_{i}^{t}$
            \EndIf
            \If{$f(pbest_i)$ is better than $f(gbest)$}
            \State Update $gbest \gets pbest_i$
            \EndIf
            \State $v_{i}^{t+1} \gets \text{Construct with } T, pbest_i  \text{ and } gbest$
            \State $\textit{prompt} \gets \text{Construct with } T, v_{i}^{t}, x_{i}^{t}, \text{ and } v_{i}^{t+1}$

            \State Query the LLM with \text{prompt} to generate a new position $x'_{i}$
            \If{$x'_{i}$ satisfies constraints}
            \State Update $x_{i}^{t+1} \gets x'_{i}$
            \Else
            \State Retry up to a fixed number of times or reinitialize $x_{i}^{t+1}$ randomly if retries exceed the limit
            \EndIf
            \EndFor
            \EndFor
            \State \Return $gbest$
        \end{algorithmic}
    \end{algorithm}
\end{figure}

The core difference between LMPSO and standard PSO only lies in the generation of the next position $x_{i}^{t+1}$ in line~15, but it fundamentally shifts how solutions are updated within PSO, making the mothod more flexible and adaptable to a wider range of problems.

The algorithm complexity of LMPSO can be represented as $O(N \cdot G \cdot C)$, where $N$ is the number of particles, $G$ is the maximum number of iterations, and $C$ is the complexity of the LLM inference process for generating a new solution.

\subsection{Meta-Prompt}
An interactive LLM is a large language model designed to facilitate natural, multi-turn conversations\cite{ouyang2022training, thoppilan2022lamda}. In this setup, the system defines the overall interaction policy and response style, the user provides instructions or queries, and the assistant (LLM) generates relevant responses.
In LMPSO, each particle acts as the “assistant”, receiving optimization instructions and producing a corresponding solution.
To enable this process, we construct a structured prompt called a \textit{meta-prompt}.
The original concept of the meta-prompt was introduced by Yang et al.\cite{yang2024large}, where it serves as a guiding template for the LLM to generate solutions.
In LMPSO, the meta-prompt is designed to follow the PSO paradigm, incorporating the following components:
\begin{enumerate}
    \item \textbf{Description of the Optimization Problem}\\
          A brief description is provided to the system, indicating the nature of the optimization task and the expected direction for generating solutions.

    \item \textbf{Inertia Term $v_{i}^{t}$}\\
          By indicating how the current position was generated (i.e., its velocity), we capture the concept of inertia. This is specified as part of the user’s instruction.

    \item \textbf{Current Position $x_{i}^{t}$}\\
          The current position serves as a reference for producing the next position. This is conveyed as part of the assistant’s response.

    \item \textbf{Direction for Generating the Next Position $v_{i}^{t+1}$}\\
          The velocity here incorporates information about the personal best and global best, thereby guiding the generation of the next position. This directive is provided as part of the user’s instruction.
\end{enumerate}
As solutions are updated, the meta-prompt is also revised, thereby adjusting the guidance each particle receives in its search for the optimal solution.

\begin{figure*}[!t]
    \centering
    \includegraphics[width=\linewidth]{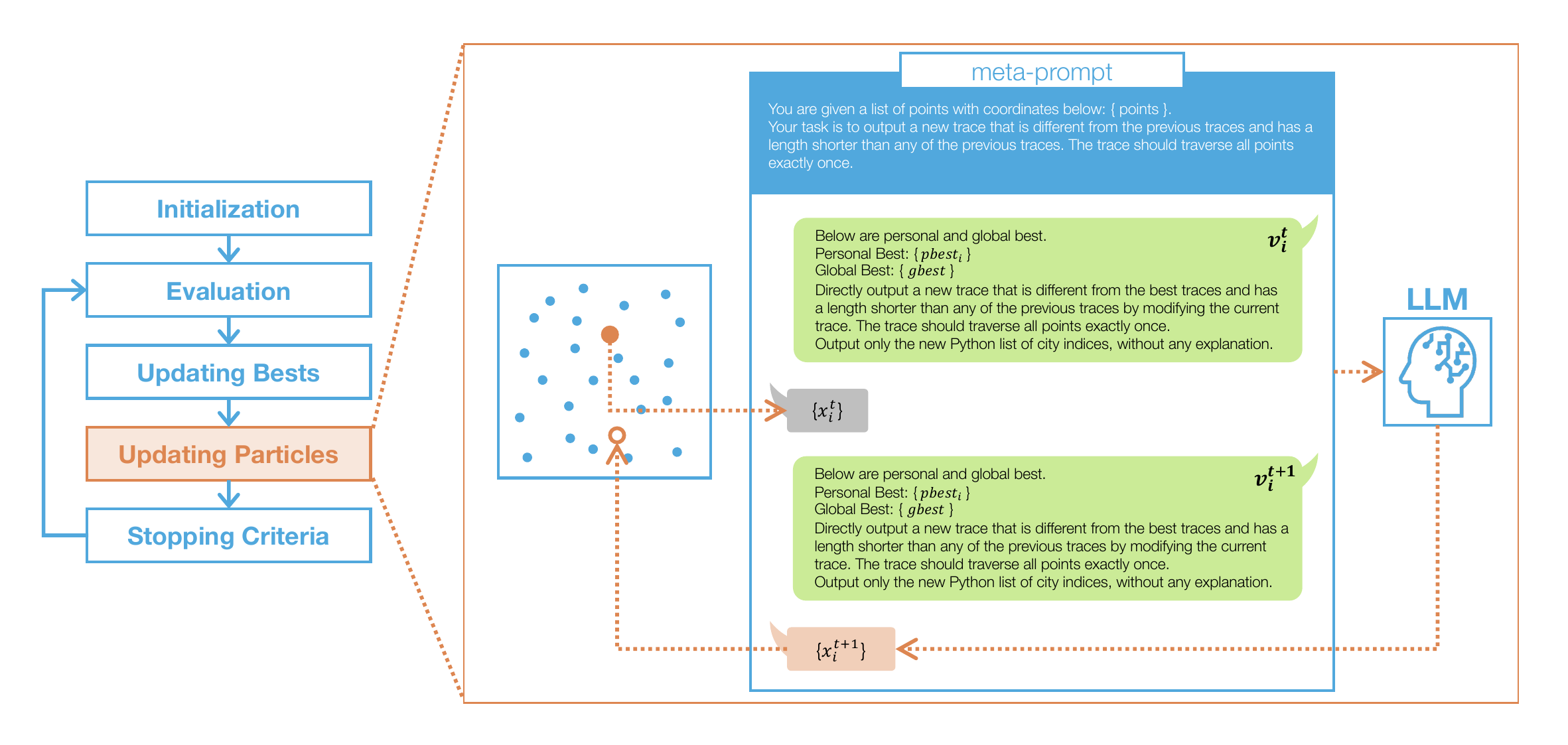}
    \caption{An overview of LMPSO. The meta-prompt shown in the figure is an example for the Traveling Salesman Problem. The components in \{\} represent the dynamic parts of the prompt which are updated according to the current state of the algorithm.}
    \label{fig::lmpso}
\end{figure*}

Fig.~\ref{fig::lmpso} illustrates an example of the meta-prompt for the Traveling Salesman Problem (TSP).
In this example, the coordinates of the cities are provided as the system content as shown in the blue box in the Fig.~\ref{fig::lmpso}.
The user's instruction is represented as the green chat bubble and the assistant's response is shown in the grey chat bubble.
The LLM generates the next position based on the interaction between the user and the assistant, which is represented as the orange chat bubble.

\section{Experiments}

We used \texttt{meta-llama/Llama-3.1-8B-Instruct} as the Large Language Model (LLM) for solution generation in this study. Llama is an open-source LLM developed by Meta, and Llama-3.1-8B-Instruct is its instruct-tuned variant~\cite{touvron2023llama}. To maintain output diversity, we set the LLM’s temperature parameter to 0.9 throughout the experiments. Because LMPSO’s runtime can grow significantly due to the inference speed of the LLM, we chose suitable values for the maximum number of iterations and the swarm size for each problem to avoid excessively long execution times.

\subsection{Combinatorial Optimization}

\subsubsection{Experimental Setup}
We evaluated the performance of our method on the Traveling Salesman Problem (TSP), in which each city’s location is given, and the goal is to determine a route that visits every city exactly once with minimal total distance. We tested three instances of TSP with 10, 20, and 30 cities, where each city’s coordinates were generated randomly in a two-dimensional plane using integer values in the range of 0--100.

We compared LMPSO against four heuristic algorithms commonly used for TSP: Nearest Neighbor (NN), Nearest Insertion (NI), Farthest Insertion (FI), and Random Insertion (RI). Each heuristic generates a candidate solution as follows:
\begin{itemize}
    \item \textbf{Nearest Neighbor (NN):} Select, among the unvisited cities, the one closest to the current city.
    \item \textbf{Nearest Insertion (NI), Farthest Insertion (FI), Random Insertion (RI):} From the unvisited cities, choose the city that is, respectively, the closest, farthest, or a random choice relative to one or more of the cities in the current tour, then insert it at the position in the route that incurs the least increase in total distance.
\end{itemize}
In addition, we compared LMPSO to a PSO tailored for TSP~\cite{wang2003particle}, which encodes particle velocity as a set of swap operations and updates positions by applying these operations.

Because generating a single solution via the LLM is computationally expensive, we set the maximum number of iterations to 100 and used 10 particles for LMPSO as shown in Table~\ref{tb::settings}.
For a fair comparison (i.e., matching the total number of objective function evaluations), the designed PSO also employed 10 particles and 100 iterations.
We conducted experiments using 5 different random city layouts for each problem size (10, 20, and 30 cities).

\begin{table*}[!t]
    \caption{Experimental settings and time costs for solving each problem using LMPSO. Time cost is reported as the mean and standard deviation over multiple runs, with the number of runs shown in parentheses.}
    \label{tb::settings}
    \begin{center}
        \begin{tabular}{|c|c|c|c|c|c|}
            \hline
            \textbf{Problem}             & \textbf{Maximum Iterations} & \textbf{Swarm Size} & \textbf{Total Evaluations} & \textbf{Max New Tokens} & \textbf{Cost (s)}   \\
            \hline
            TSP (10 cities)              & 100                         & 10                  & 1,000                      & 50                      & 681.0 $\pm$ 7.0 (5) \\
            TSP (20 cities)              & 100                         & 10                  & 1,000                      & 100                     & 1411 $\pm$ 26 (5)   \\
            TSP (30 cities)              & 100                         & 10                  & 1,000                      & 150                     & 3340 $\pm$ 150 (5)  \\
            Heuristic Improvement        & 40                          & 25                  & 1,000                      & 1000                    & 34,900 $\pm$ 0 (1)  \\
            Symbolic Regression (dim-2)  & 50                          & 80                  & 4,000                      & 200                     & 5810 $\pm$ 970 (5)  \\
            Symbolic Regression (dim-5)  & 50                          & 80                  & 4,000                      & 200                     & 5830 $\pm$ 540 (5)  \\
            Symbolic Regression (dim-10) & 50                          & 80                  & 4,000                      & 200                     & 5760 $\pm$ 680 (5)  \\
            \hline
        \end{tabular}
    \end{center}
\end{table*}

\subsubsection{Results}
Table~\ref{tb::tsp} summarizes the results for TSP with 10, 20, and 30 cities. Each algorithm’s performance was evaluated using five different random city layouts, and we report the mean and standard deviation of the Optimality Gap relative to the global optimum (computed by the Concorde TSP solver~\cite{applegate2006concorde}).
\begin{table}[!t]
    \caption{Results on TSP instances.}
    \begin{center}
        \begin{tabular}{|c|c|c|c|}
            \hline
            \textbf{}       & \multicolumn{3}{|c|}{\textbf{Optimality Gap (\%)}}                                                       \\
            \cline{2-4}
            \textbf{Method} & \textbf{10 cities}                                 & \textbf{20 cities}       & \textbf{30 cities}       \\
            \hline
            NN              & $0.11 \pm 0.09$                                    & $0.14 \pm 0.05$          & $0.20 \pm 0.07$          \\
            NI              & $0.06 \pm 0.04$                                    & $0.14 \pm 0.06$          & $0.16 \pm 0.06$          \\
            FI              & $\mathbf{0.00 \pm 0.00}$                           & $\mathbf{0.01 \pm 0.01}$ & $\mathbf{0.05 \pm 0.03}$ \\
            RI              & $0.85 \pm 0.08$                                    & $1.74 \pm 0.17$          & $2.25 \pm 0.36$          \\
            PSO             & $0.08 \pm 0.08$                                    & $0.90 \pm 0.07$          & $1.39 \pm 0.07$          \\
            LMPSO           & $0.02 \pm 0.02$                                    & $0.42 \pm 0.12$          & $0.73 \pm 0.05$          \\
            \hline
        \end{tabular}
        \label{tb::tsp}
    \end{center}
\end{table}

As shown in Table~\ref{tb::tsp}, LMPSO achieves competitive performance on the 10-city instance, ranking second only to FI. Notably, it outperforms the TSP-specific PSO in all problem sizes, suggesting that LMPSO is effective even when applied to combinatorial optimization problems. However, for 20-city and 30-city instances, LMPSO’s performance is lower than that of heuristics such as NN, NI, and FI.

The time taken for each problem is shown in Table~\ref{tb::settings}.
The time taken for the problem increases as the number of cities increases, with the 30-city instance requiring the most time.

\subsection{Heuristic Improvement}

One advantage of using an LLM is the ability to treat natural language as the search space. In fact, many studies have attempted to generate better programs using LLMs~\cite{chen2024evoprompting,Romera-Paredes:2024aa}.
In this experiment, we investigated whether LMPSO could improve existing TSP heuristic algorithms.

\subsubsection{Experimental Setup}
We treated the Python implementations of NN, NI, FI, and RI as strings representing the heuristics. We randomly chose one of these four as the initial solution for each particle and then ran LMPSO to improve the heuristic. We tested 5 different 100-city TSP instances and used the total distance obtained from each heuristics as the objective function value. After exploring several combinations of the maximum number of iterations and swarm sizes, we found that 40 iterations and 25 particles (i.e., 1,000 total solutions) were sufficient to provide good performance while considering time constraints. We chose 1,000 total solutions to ensure a wide variety of heuristics could be generated without excessively long run times.

\subsubsection{Heuristic Generated by LMPSO}
By running LMPSO on the Python implementations of NN, NI, FI, and RI (treated as string-based heuristics), we obtained a hybrid heuristic that combines several decision criteria and random operations.
The resulting algorithm operates as follows:
\begin{enumerate}
    \item Create \texttt{tour}, initially containing only city 0, and place the remaining cities into \texttt{remaining}.
    \item Repeatedly insert unvisited cities into the tour, using the following decision patterns:
          \begin{itemize}
              \item \textbf{Pattern A}: If the $x$-coordinate of the last city in the tour is greater than the global minimum $x$-value:
                    \begin{itemize}
                        \item \texttt{farthest} = the city among unvisited ones whose distance to any city in the tour is maximal.
                        \item \texttt{farthest\_from\_center} = in practice, the city closest to the tour center (\texttt{tour\_center}).
                        \item Determine which city to insert based on angles and distance to the center.
                    \end{itemize}
              \item \textbf{Pattern B}: Otherwise (i.e., if the last city’s $x$-coordinate is not larger than the global minimum $x$-value):
                    \begin{itemize}
                        \item \texttt{nearest\_to\_median} = the city closest to the center.
                        \item \texttt{farthest\_from\_median} = in practice, the city whose distance from the last city is maximal.
                        \item Decide which city to insert by comparing distances to the center.
                    \end{itemize}
          \end{itemize}
    \item Insert the chosen city at the position in the tour that minimizes additional travel distance.
    \item Perform random partial reversals (\texttt{tour[::-1]}) or partial rotations (moving some leading or trailing cities) with a certain probability to avoid local optima.
    \item Repeat these steps until all cities have been inserted into the tour.
\end{enumerate}

Starting from the existing heuristics (NN, NI, FI, RI), LMPSO produced a new heuristic that mixes prior elements with new elements such as angles, center-based calculations, random reversals and partial rotations. While this resulted in a unique approach, the generated code sometimes featured inconsistencies between variable names and their actual behavior, as well as potentially redundant or random operations whose direct effect on overall performance remained unclear.

\subsubsection{Search Process}

\begin{figure}
    \centering
    \includegraphics[width=0.9\linewidth]{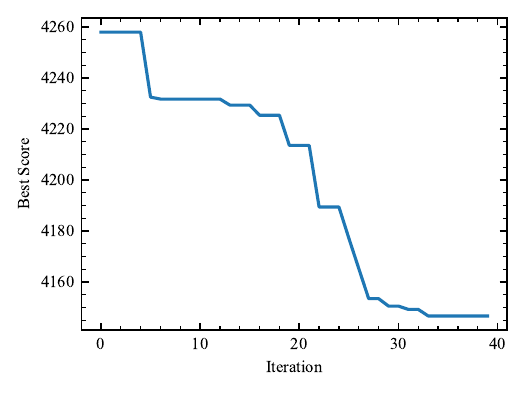}
    \caption{Search process of LMPSO for heuristic improvement on the TSP. The vertical axis represents the total travel distance obtained by applying the current best heuristic to five distinct 100-city TSP instances.}
    \label{fig::tsp_search}
\end{figure}

Fig.~\ref{fig::tsp_search} illustrates how LMPSO explores and refines heuristics over multiple iterations.
Solutions generated early in the search often featured intricate branching conditions, insertion strategies, or elements such as a “center” or “density.”
However, except for the significant improvement in iteration~5, these initial approaches failed to improve the heuristic gradually, as shown in Fig.~\ref{fig::tsp_search}.

During the mid-phase of the search, LMPSO continued to refine branching conditions and methods for determining insertion locations.
Notably, angle-based criteria began appearing in some solutions around iteration~21.
By approximately iteration~25, techniques such as partial reversals, tour rotations, and angle-based strategies emerged in the best-performing solutions.
These techniques persisted into the final stages of the search, which involved further refinement of branching criteria, insertion methods, and parameter tuning.

Ultimately, LMPSO identified its best solution at iteration~33.
The improvements leading to this point included detailed angle-based logic and route-modification operations, which had gradually evolved from the foundational strategies of earlier iterations.

\subsubsection{Comparison with Other Methods}
Table~\ref{tb::tsp_improvement} shows the results of applying the heuristics generated by LMPSO and other baseline heuristics to five 100-city TSP instances. The Optimality Gap values represent the mean and standard deviation across these five test instances.
\begin{table}[!t]
    \caption{Results of Improving Heuristics.}
    \begin{center}
        \begin{tabular}{|c|c|}
            \hline
            \textbf{}       & \textbf{Optimality Gap (\%)} \\
            \cline{2-2}
            \textbf{Method} & \textbf{100 cities}          \\
            \hline
            NN              & $0.28 \pm 0.09$              \\
            NI              & $0.21 \pm 0.04$              \\
            FI              & $0.07 \pm 0.01$              \\
            RI              & $5.69 \pm 0.37$              \\
            PSO             & $4.60 \pm 0.23$              \\
            LMPSO           & $\mathbf{0.06 \pm 0.02}$     \\
            \hline
        \end{tabular}
        \label{tb::tsp_improvement}
    \end{center}
\end{table}
As shown in Table~\ref{tb::tsp_improvement}, the LMPSO-generated heuristic achieved the best performance among all compared methods.

\subsection{Symbolic Regression}

\subsubsection{Experimental Setup}
Symbolic regression aims to discover an optimal mathematical expression that fits a given set of data points.
In the experiment of the combinatorial optimization problem, the solutions generated by the LLM are a sequence of numbers that hold little inherent meaning on their own.
However, in symbolic regression, the solutions are mathematical expressions that have a clear structure and meaning.
By testing LMPSO on symbolic regression tasks, we tried to explore the method’s effectiveness in solving problems with structured solutions.

For LMPSO, the meta-prompt included a description of the symbolic regression task, a random subset of 20 data points, instructions to produce diverse expressions, and a note that shorter expressions were preferable.
The random subset of 20 data points were included in the prompt to guide the LLM in generating expressions that fit the data.
The initial solutions were expressions generated by prompting the LLM with the same problem description and data points.

Genetic Programming (GP) is widely used for solving this task~\cite{augusto2000symbolic}, as it represents expressions as trees that evolve over time.
We compared LMPSO to the \texttt{gplearn} GP library on symbolic regression tasks from the Penn Machine Learning Benchmarks (PMLB)~\cite{romano2021pmlb}, focusing on black-box functions ranging from 2 to 10 dimensions.
Because the true functional form is unknown, we measured performance using the Mean Absolute Error (MAE).

After testing various combinations of the maximum number of iterations and swarm sizes (to achieve 4,000 total evaluations), we found that 50 iterations and 80 particles provided a reasonable balance between solution diversity and convergence.
For a fair comparison, we also set GP’s maximum iterations to 50 and its population size to 80, resulting in the same total number of evaluations as LMPSO.
We adopted the typical GP objective function (MAE) and included the following operators: addition, subtraction, multiplication, division, log, sqrt, abs, neg, inv, max, and min.
In the meta-prompt of LMPSO, we also included instructions to use these operators if necessary to fit the data.
Finally, we set the \texttt{gplearn} parameters to a crossover probability of 0.7 and a mutation probability of 0.1.
We conducted five runs for each problem instance to evaluate the performance of each method.

\subsubsection{Comparison of Results}
We compared LMPSO and GP on symbolic regression tasks from PMLB (2D--10D). Both methods were run five times, and we report the mean and standard deviation of the coefficient of determination ($R^2$) for the final solutions in Table~\ref{tb::symbolic_regression}. A value of $R^2$ close to 1 indicates high predictive accuracy.
\begin{table}[!t]
    \caption{Results for Symbolic Regression Tasks.}
    \begin{center}
        \begin{tabular}{|c|c|c|c|}
            \hline
            \textbf{}             & \textbf{}    & \multicolumn{2}{|c|}{R\textsuperscript{2}}                    \\
            \cline{3-4}
            \textbf{Dataset Name} & \textbf{Dim} & \textbf{LMPSO}                             & \textbf{GP}      \\
            \hline
            vineyard              & 2            & $\mathbf{0.68 \pm 0.05}$                   & $0.42 \pm 0.11$  \\
            analcatdata\_apnea2   & 3            & $-0.07 \pm 0.00$                           & $-0.07 \pm 0.00$ \\
            ESL                   & 4            & $\mathbf{0.78 \pm 0.02}$                   & $0.70 \pm 0.02$  \\
            cloud                 & 5            & $\mathbf{0.85 \pm 0.02}$                   & $0.71 \pm 0.03$  \\
            machine\_cpu          & 6            & $\mathbf{0.87 \pm 0.06}$                   & $0.61 \pm 0.17$  \\
            pm10                  & 7            & $\mathbf{0.16 \pm 0.03}$                   & $-0.13 \pm 0.14$ \\
            house\_8L             & 8            & $\mathbf{0.03 \pm 0.16}$                   & $-1.33 \pm 0.69$ \\
            BNG\_lowbwt           & 9            & $\mathbf{0.06 \pm 0.50}$                   & $-0.62 \pm 0.61$ \\
            SWD                   & 10           & $\mathbf{0.26 \pm 0.05}$                   & $-0.17 \pm 0.21$ \\
            \hline
        \end{tabular}
        \label{tb::symbolic_regression}
    \end{center}
\end{table}
From Table~\ref{tb::symbolic_regression}, LMPSO produced expressions with higher $R^2$ values than those from GP, indicating superior modeling ability on most test problems.

\subsubsection{Search Process}

\begin{figure}[!t]
    \centering
    \begin{minipage}{0.49\textwidth}
        % \centering
        % \includegraphics[width=\linewidth]{figures/best_score_8_points.pdf}
        \centerline{\includegraphics{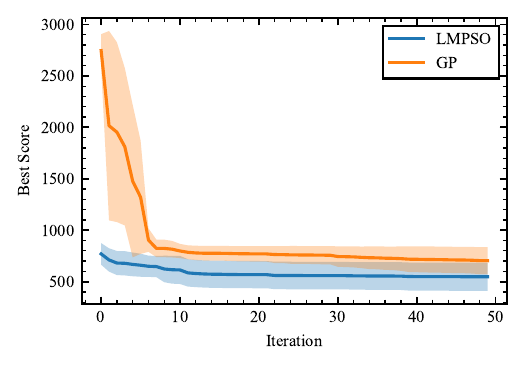}}
        \subcaption{MAE of the Best Solutions}
    \end{minipage}
    \hspace{0.02\textwidth}
    \begin{minipage}{0.49\textwidth}
        % \centering
        % \includegraphics[width=\linewidth]{figures/best_length_8_points.pdf}
        \centerline{\includegraphics{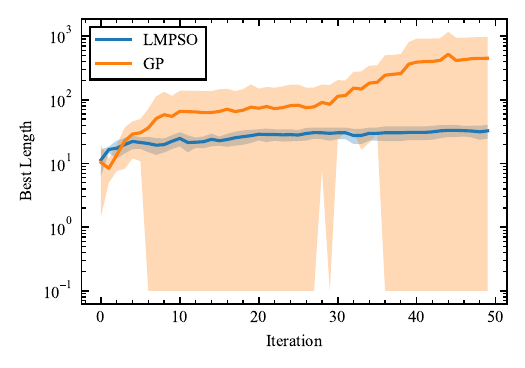}}
        \subcaption{Expression Length of the Best Solutions}
    \end{minipage}
    \caption{Comparison of LMPSO and GP over five runs for \texttt{1193\_BNG\_lowbwt} (9D). “Best Score” represents the Mean Absolute Error of the cumulative best solution, while “Best Length” indicates the length of the best solution at each iteration. The shaded area represents the standard deviation, with a minimum of 0.1 set to accommodate logarithmic scaling in (b).}
    \label{fig::symbolic_regression}
\end{figure}

We analyzed the results for the 9D dataset \texttt{1193\_BNG\_lowbwt}, focusing on the best MAE and expression length during the optimization process.
Fig.\ref{fig::symbolic_regression} compares the performance of LMPSO and GP throughout the optimization.
Since a lower MAE indicates a better solution, LMPSO consistently outperformed GP from the initial iterations and maintained a lower MAE throughout the search, as shown in Fig.\ref{fig::symbolic_regression}(a).
Shorter expressions are generally associated with better generalization in symbolic regression. The best solutions generated by GP were significantly longer than those produced by LMPSO, highlighting LMPSO’s advantages in this regard (Fig.\ref{fig::symbolic_regression}(b)).
Furthermore, Fig.\ref{fig::symbolic_regression}(b) shows that the length of the best solutions generated by LMPSO remained relatively stable across iterations and different runs, demonstrating LMPSO’s ability to consistently produce concise solutions with shorter expression lengths.

\subsubsection{Generated Solutions}

Table~\ref{tb::symbolic_regression_2d} presents the solutions generated by LMPSO for the 2D dataset \texttt{vineyard}.
As shown in Table~\ref{tb::symbolic_regression_2d}, iteration~1 produces a very simple expression that employs  $\lvert x_1 - 10 \rvert$. Subsequently, around iterations~5 to 10, the model begins incorporating squared terms such as $(x_1 - 11)^2$ to further adjust the error with respect to the data (for instance, $\frac{(x_1 - 11)^2}{5}$ appears at iteration~10). Between iterations~15 and 30, additional terms related to $x_0$, such as $(x_0 - 3)$ and $(x_0 - 4)^2$, are introduced, reflecting interactions among multiple variables and indicating further parameter fine-tuning. Moreover, starting from iteration~35, $\sin$ terms are introduced to account for periodic variations and capture subtle changes in the data. In fact, by iteration~50, the algorithm yields a more advanced expression combining multiple $\sin$ terms (e.g., $\sin(x_1 - 10.51)$, $\sin(x_1 - 10.54)$), demonstrating that LMPSO thoroughly explores the solution space while increasing the complexity of the model.
The tendency to generate simple expression at the beginning and gradually increase complexity is consistent with the behavior observed in the search process for symbolic regression tasks.

\begin{table*}[!t]
    \caption{Best Solutions for the Iteration of LMPSO on the 2D Dataset \texttt{vineyard}.}
    \begin{center}
        \begin{tabular}{|c|l|}
            \hline
            \textbf{Iteration} & \textbf{Best Solution}                                                                                                                           \\ \hline
            1                  & $20 - 2 \cdot \lvert x_1 - 10 \rvert$                                                                                                            \\
            5                  & $20 + (x_0 - 3) - (\lvert x_1 - 10 \rvert + 0.5)$                                                                                                \\
            10                 & $20 + (x_0 - 3) - \left(\frac{(x_1 - 11)^2}{5} + 0.1\right)$                                                                                     \\
            15                 & $20 + (x_0 - 3) - 0.4 \cdot \left(\frac{(x_1 - 11)^2}{5} + (x_0 - 4)^2\right)$                                                                   \\
            20                 & $20 + (x_0 - 3) - 0.46 \cdot \left(\frac{(x_1 - 11)^2}{5} + (x_0 - 4)^2\right) + \frac{x_1 - 10}{100}$                                           \\
            25                 & $20 + (x_0 - 3) - 0.457 \cdot \left(\frac{(x_1 - 11)^2}{5} + (x_0 - 4)^2\right) + 0.025 \cdot (x_1 - 10)$                                        \\
            30                 & $20 + (x_0 - 3) - 0.455 \cdot \left(\frac{(x_1 - 11)^2}{5} + (x_0 - 4)^2\right) + 0.05 \cdot (x_1 - 10)$                                         \\
            35                 & $20 + (x_0 - 3) - 0.45 \cdot \left(\frac{(x_1 - 11)^2}{5} + (x_0 - 4)^2\right) + 0.05 \cdot (x_1 - 10 + \sin(x_1 - 10))$                         \\
            40                 & $20 + (x_0 - 3) - 0.45 \cdot \left(\frac{(x_1 - 11)^2}{5} + (x_0 - 4)^2\right) + 0.525 \cdot \sin(x_1 - 11)$                                     \\
            45                 & $20 + (x_0 - 3) - 0.45 \cdot \left(\frac{(x_1 - 11)^2}{5} + (x_0 - 4)^2\right) + 0.55 \cdot (\sin(x_1 - 10.5) + 0.1 \cdot \sin(x_1 - 10.55))$    \\
            50                 & $20 + (x_0 - 3) - 0.446 \cdot \left(\frac{(x_1 - 11)^2}{5} + (x_0 - 4)^2\right) + 0.576 \cdot \sin(x_1 - 10.51) + 0.084 \cdot \sin(x_1 - 10.54)$ \\ \hline
        \end{tabular}
        \label{tb::symbolic_regression_2d}
    \end{center}
\end{table*}

\section{Discussions and Conclusion}
In this work, we proposed \textit{LMPSO}, an optimization method that directly employs an interactive Large Language Model (LLM) to generate solutions within the Particle Swarm Optimization (PSO) framework. We applied LMPSO to three types of problems: the Traveling Salesman Problem (TSP), heuristic-improvement tasks (where solutions are treated as strings in a natural-language search space), and symbolic regression.

Based on our experimental results, we highlight several key observations regarding LMPSO:

\begin{itemize}
    \item \textbf{An Extension of PSO Using LLMs.}
          While traditional PSO often requires problem-specific designs (e.g., tailoring velocity updates, solution representations), LMPSO preserves the PSO framework but relies on prompt engineering to adapt to various tasks. In other words, \emph{simply changing the prompt} allows LMPSO to tackle different optimization problems, including those involving natural-language solution representations, which would be impractical in standard PSO.

    \item \textbf{Challenges for Large-Scale Problems.}
          Similar to prior research, using an LLM for direct optimization of large-scale solution representations proved difficult. However, our experiments suggest that converting combinatorial tasks into heuristic-improvement problems enables LMPSO to effectively generate solutions without manipulating extensive solution representations directly.

    \item \textbf{Effectiveness as a Hyper-Heuristic.}
          When applied to TSP heuristics, LMPSO demonstrated its effectiveness as a hyper-heuristic by generating high-performing hybrid heuristics. By combining elements from multiple existing heuristics with novel components (e.g., center- or angle-based operations) that are absent in the traditional methods, LMPSO showcased its capability to extend beyond human-designed approaches and create innovative solutions.

    \item \textbf{Performance in Symbolic Regression.}
          In symbolic regression experiments, LMPSO achieved higher coefficients of determination than a standard Genetic Programming (GP) library. By including prompts encouraging shorter expressions, LMPSO was able to generate concise yet accurate solutions.

\end{itemize}

\textbf{Future Directions.} Several issues remain for further investigation:
\begin{itemize}
    \item \textbf{Extending LMPSO to Other Problem Domains.}
          As LMPSO can operate on solutions represented in natural language, evaluating its performance across a broader range of problems would be valuable.

    \item \textbf{Assessing the Impact of LLM Performance.}
          Because LMPSO’s efficacy may depend heavily on the capabilities of the underlying LLM, quantifying how different LLM architectures or parameter sizes affect LMPSO’s performance is an important research avenue.

    \item \textbf{Incorporating PSO Enhancements.}
          LMPSO follows the PSO framework and can therefore benefit from various enhancements proposed for PSO. Investigating how these enhancements interact with LMPSO could lead to further improvements in solution quality and convergence speed.

    \item \textbf{Refining LLM Utilization.}
          Advanced prompting methods such as “Chain of Thought”~\cite{wei2022chain} could more effectively exploit an LLM’s reasoning abilities, and introducing multi-agent systems for particle-level communication may further boost performance~\cite{chan2023chateval}.
\end{itemize}

In conclusion, LMPSO demonstrates the potential of integrating Large Language Models into the PSO framework to address a wide range of optimization tasks.
By leveraging natural-language prompts instead of specialized operators, LMPSO lowers the barrier to applying PSO to diverse problems.
We believe that LMPSO not only broadens the scope of swarm intelligence methods but also opens up new possibilities for leveraging LLMs in optimization tasks.

\section*{Acknowledgment}

We acknowledge the use of ChatGPT-4o for polishing and improving the clarity of the text across this paper. This assistance was limited to text refinement and did not influence the content, structure, or conclusions of the work.
\bibliographystyle{IEEEtran}
\bibliography{IEEEabrv, ref}

\end{document}